\documentclass[lettersize,journal]{IEEEtran}
\usepackage{array}
\usepackage{textcomp}
\usepackage{stfloats}
\usepackage{verbatim}

\usepackage{amsmath,amssymb,amsfonts}
\usepackage{algpseudocode,algorithm}
\usepackage{subcaption}
\usepackage[T1]{fontenc}

\usepackage{graphicx}
\usepackage{cite}
\usepackage{url}
\begin{document}

\title{Soft Actor-Critic Algorithm with \\Truly-satisfied Inequality Constraint}

\author{Taisuke Kobayashi,~\IEEEmembership{Member,~IEEE}
        % <-this % stops a space
\thanks{Taisuke Kobayashi is with National Institute of Informatics (NII) and with The Graduate University for Advanced Studies (SOKENDAI), 2-1-2 Hitotsubashi, Chiyoda-ku, Tokyo, 101-8430, Japan.}% <-this % stops a space
\thanks{Manuscript received April 19, 2021; revised August 16, 2021.}}

\markboth{Journal of \LaTeX\ Class Files,~Vol.~14, No.~8, August~2021}%
{Shell \MakeLowercase{\textit{et al.}}: Soft Actor-Critic Algorithm with Truly-satisfied Inequality Constraint}

\IEEEpubid{0000--0000/00\$00.00~\copyright~2021 IEEE}

\maketitle

%%%%%%%%%%%%%%%%%%%%%%%%%%%%%%%%%%%%%%%%%%%%%%%%%%%%%%%%%%%%%%%%%%%%%%%%%%%%%%%%
% no more than 250 words
\begin{abstract}

Soft actor-critic (SAC) in reinforcement learning is expected to be one of the next-generation robot control schemes.
Its ability to maximize policy entropy would make a robotic controller robust to noise and perturbation, which is useful for real-world robot applications.
However, the priority of maximizing the policy entropy is automatically tuned in the current implementation, the rule of which can be interpreted as one for equality constraint, binding the policy entropy into its specified lower bound.
The current SAC is therefore no longer maximize the policy entropy, contrary to our expectation.
To resolve this issue in SAC, this paper improves its implementation with a learnable state-dependent slack variable for appropriately handling the inequality constraint to maximize the policy entropy by reformulating it as the corresponding equality constraint.
The introduced slack variable is optimized by a switching-type loss function that takes into account the dual objectives of satisfying the equality constraint and checking the lower bound.
In Mujoco and Pybullet simulators, the modified SAC statistically achieved the higher robustness for adversarial attacks than before while regularizing the norm of action.
A real-robot variable impedance task was demonstrated for showing the applicability of the modified SAC to real-world robot control.
In particular, the modified SAC maintained adaptive behaviors for physical human-robot interaction, which had no experience at all during training.

\end{abstract}

%%%%%%%%%%%%%%%%%%%%%%%%%%%%%%%%%%%%%%%%%%%%%%%%%%%%%%%%%%%%%%%%%%%%%%%%%%%%%%%%
%%%%%%%%%%%%%%%%%%%%%%%%%%%%%%%%%%%%%%%%%%%%%%%%%%%%%%%%%%%%%%%%%%%%%%%%%%%%%%%%
\section{Introduction}

%%%robot control w/ rl
\IEEEPARstart{F}{or} the accomplishment of increasingly complex robot tasks (e.g. deformable object manipulation~\cite{tsurumine2019deep} and human-robot interaction~\cite{kobayashi2021whole}), data-driven control, especially (deep) reinforcement learning (RL)~\cite{sutton2018reinforcement}, is expected to be next-generation control scheme.
Although learning had been limited to game environments and dynamics simulators, it has recently begun to be applied to real-world robot control and has illustrated outcomes~\cite{andrychowicz2020learning,osinski2020simulation,li2021reinforcement}.

%%%sac
One of the key players in this success is the state-of-the-art learning algorithm, soft actor-critic (SAC)~\cite{haarnoja2018soft}.
As its name suggests, SAC is a type of actor-critic algorithm, which is suitable for robot control because it allows control commands (a.k.a. actions in RL) given in real numbers, such as joint torques, by directly optimizing a stochastic policy to generate them.
SAC renewed the learning law by adding maximization of policy entropy into the main objective of RL, which facilitates exploration while suppressing over-learning and improves learning quality and speed.
In addition, as a side effect of the maximized policy entropy in its learning process, the policy can acquire the superior ability to recover from errors in various situations, which are encountered during the encouraged exploration~\cite{haarnoja2018soft2,eysenbach2021maximum}.
This ability corresponds to the robustness to noise and perturbation, which should be required for the real-world robot control as well as the control quality.
Furthermore, various improvements have been proposed using SAC as a baseline:
e.g. making SAC suitable for discrete action space~\cite{christodoulou2019soft};
improving the policy model by combining normalizing flows~\cite{ward2019improving} and;
ensuring safety by considering the worst-case scenario~\cite{yang2021wcsac}.
Alternatively, SAC has been re-derived and analyzed in the context of control as inference~\cite{levine2018reinforcement}.

\IEEEpubidadjcol
%%%auto-tuned temperature
One direction of improving SAC is the auto-tuning of the temperature parameter that determines the priority of the maximization of the policy entropy, which is added to SAC as an auxiliary objective.
This function eliminates the need for hand-tuning according to tasks.
The most representative auto-tuning was developed by the authors who originally proposed SAC~\cite{haarnoja2018soft2}.
By reinterpreting SAC as a constrained optimization problem once, the temperature parameter is given the meaning of a Lagrange multiplier, optimization procedure of which can be applied to the temperature parameter.
Popular libraries have implemented SAC with this function~\cite{liang2018rllib,raffin2021stable}, and empirically proven its simple and effective effects, although its issue is raised later.
Other methods have been studied, such as the use of meta-gradient~\cite{wang2020meta} and the introduction of curiosity to further accelerate the exploration~\cite{lin2020cat}, but their improvement seems to be marginal.

%%%target
Thus, SAC has been established as a modern standard RL algorithm, but as mentioned above, it is still in the process of improvement.
In this study, the auto-tuning of the temperature parameter is modified.
Specifically, this paper first points out that the current standard implementation can be regarded as a derivation with an equality constraint on the policy entropy, which causes the policy entropy to converge to its lower bound without maximizing it~\cite{xu2021target}.
In other words, the current auto-tuning of the temperature parameter would make SAC far from our expectation to maximize the policy entropy.
If the policy entropy on SAC is only to be kept constant, it can be fixed at the design stage, such as in DDPG~\cite{lillicrap2016continuous} and TD3~\cite{fujimoto2018addressing}, and there might be little benefit from the regularization of the policy entropy.
Unless the ability to flexibly adjust the policy entropy above the lower bound, which SAC is supposed to possess, is properly incorporated, it is difficult to truly find the benefits of SAC.

%%%proposal
To resolve this issue, this paper introduces and optimizes a learnable state-dependent slack variable for inequality constraint~\cite{boyd2004convex} to maximize the policy entropy so that it can appropriately be over a specified lower bound.
The conventional auto-tuning can correctly satisfy the inequality constraint of the policy entropy by simply adding the slack variable to the lower bound.
In addition, as the slack variable is approximated as a function with state input using deep neural networks, it can be easily optimized through minimizing a loss function that reflects its requirements (i.e. proper transformation to the equality constraint and verification of the lower bound).
For such a multi-objective optimization, a switching-type scalarization instead of a simple weighted sum is introduced based on $\epsilon$-insensitive loss~\cite{vapnik1999nature}.
This improvement is expected to appropriately yield the policy entropy maximization with the high robustness and alleviation of over-learning for more stability in the practical use.

%%%results
The proposed method is statistically validated on Mujoco~\cite{todorov2012mujoco} and Pybullet~\cite{coumans2016pybullet} simulators.
The results show that the modified SAC with the slack-variable-used auto-tuning adjusts (mainly increases) the policy entropy.
Such an entropy-maximized policy yields the higher robustness to adversarial situations while conservatively regularizing the norm of action.
In addition, a real-robot variable impedance task is demonstrated for showing the applicability of the proposed method to real-world robot control.
The modified SAC exhibits adaptive behaviors even for physical human-robot interaction, which had no experience at all during training.

%%%%%%%%%%%%%%%%%%%%%%%%%%%%%%%%%%%%%%%%%%%%%%%%%%%%%%%%%%%%%%%%%%%%%%%%%%%%%%%%
\section{Preliminaries}

%%%%%%%%%%%%%%%%%%%%%%%%%%%%%%%%%%%%%%%%%%%%%%%%%%%%%%%
\subsection{Basic problem statement}

RL~\cite{sutton2018reinforcement} solves an optimal control problem that maximizes the sum of future rewards, so-called return.
For this purpose, Markov decision process (MDP) is assumed with the tuple $(\mathcal{S}, \mathcal{A}, p_e, r)$ where $\mathcal{S}$ and $\mathcal{A}$ denote the state and action spaces, respectively, $p_e$ gives the state transition probability, and $r$ defines the reward at the current situation.
Specifically, at the current state $s \in \mathcal{S}$ of an (unknown) environment, an agent decides its action $a \in \mathcal{A}$ according to its trainable policy $\pi(a \mid s)$ and interacts with the environment by $a$.
As a result, $s$ is updated to the next state $s^\prime$ according to the state transition probability $p_e(s^\prime \mid s, a)$, while getting the reward $r = r(s, a)$.

RL optimizes $\pi$ to stochastically maximize the return obtained through interactions repeated from $s$ at the current time step $t$, $R_t = \sum_{k=0}^\infty \gamma^k r(s_{t+k}, a_{t+k})$ with $\gamma \in [0, 1)$ the discount factor.
For this optimization problem, RL often defines the following two types of value functions, which denote the expected values of $R_t$.
\begin{align}
    V(s) = \mathbb{E}_{\rho^\pi}[R_t \mid s]
    , \
    Q(s,a) = \mathbb{E}_{\rho^\pi}[R_t \mid s, a]
\end{align}
where $\rho^\pi$ denotes the probability to generate the state-action trajectory under $\pi$ (and $p_e$).
Note that $V(s) = \mathbb{E}_{a \sim \pi}[Q(s, a)]$ holds in this standard case.
Various learning algorithms have been proposed for learning these functions as well as $\pi$, and SAC, described below, is one of them.
Note that, to learn $\pi$, $V$ and/or $Q$, deep neural networks are widely used as function approximators of them.
This paper also follows this popular way of implementation.

%%%%%%%%%%%%%%%%%%%%%%%%%%%%%%%%%%%%%%%%%%%%%%%%%%%%%%%
\subsection{Soft actor-critic: SAC}

Let's address SAC, which has achieved great success in recent years in RL and has become established as one of the standard methods~\cite{haarnoja2018soft}.
Its main feature is that it solves the entropy-maximized problem by adding the policy entropy $\mathcal{H}(\pi) = \mathbb{E}_{\pi}[-\ln\pi]$ to the reward as follows:
\begin{align}
    r = r(s, a) + \alpha \mathcal{H}(\pi(\cdot \mid s))
\end{align}
where $\alpha \geq 0$ denotes the temperature parameter to adjust the priority of the policy entropy.

According to this extension, the relationship between $V$ and $Q$ is first given as follows:
\begin{align}
    V(s) &= \mathbb{E}_{a \sim \pi}[Q(s, a) - \alpha \ln \pi(a \mid s)]
\end{align}
With this, the soft Bellman equation derives the following bootstrapped supervised signal for $Q$ as $y$.
\begin{align}
    &Q(s, a)
    \nonumber \\
    &= r(s, a) + \gamma \mathbb{E}_{p_{e}(s^\prime \mid s, a)}[V(s^\prime)]
    \nonumber \\
    &= r(s, a) + \gamma \mathbb{E}_{p_{e}(s^\prime \mid s, a), \pi(a^\prime \mid s^\prime)}[Q(s^\prime, a^\prime) - \alpha \ln \pi(a^\prime \mid s^\prime)]
    \nonumber \\
    &= y
\end{align}
It is known that $Q$ can be correctly obtained as the expected value of the return by optimizing the approximated $Q$ in order to satisfy this equality.

In the actual implementation of SAC, two different function approximators are prepared to approximate $Q_{1,2}$, which are utilized for avoiding overestimation of $Q$.
In addition, the target networks $\bar{Q}_{1,2}$~\cite{kobayashi2022consolidated} corresponding to them are provided to compute $y$.
Therefore, learning of $Q_{1,2}$ is accomplished by solving the following minimization problem for temporal difference (TD) error.
\begin{align}
    Q_{1,2}^{\ast} &= \arg \min_{Q_{1,2}} \mathbb{E}_{(s, a, s^\prime, r) \sim \mathcal{D}, a^\prime \sim \pi}
    \left[ \frac{1}{2}(y - Q_{1,2}(s, a))^2 \right]
    \label{eq:loss_Q} \\
    y &= r + \gamma \Bigl\{ \min_{i=1,2}\bar{Q}_{i}(s^\prime, a^\prime) - \alpha \ln \pi(a^\prime \mid s^\prime) \Bigr\}
    \nonumber
\end{align}
where $\mathcal{D} = \{(s, a, s^\prime, r)_i\}_{i=1}^N$ denotes the replay buffer of size $N$, in which the interaction data on MDP is stored as the tuple $(s, a, s^\prime, r)$.

With the trained $Q$, $\pi$ can be optimized to maximize $V$ (or minimize $-V$) at any states as follows:
\begin{align}
    \pi^{\ast} &= \arg \min_{\pi} \mathbb{E}_{s \sim \mathcal{D}}[-V(s)]
    \nonumber \\
    &= \arg \min_{\pi} \mathbb{E}_{s \sim \mathcal{D}, a \sim \pi}
    \Bigl[ - \min_{i=1,2} Q_{i}(s, a) + \alpha \ln \pi(a \mid s) \Bigr]
    \label{eq:loss_pi}
\end{align}
Here, by maximizing the smaller $Q$, a conservative update would be expected.
This minimization problem can be solved using the reparameterization trick for $a \sim \pi$~\cite{kingma2014auto}.
That is, the gradients of the above loss function w.r.t. $\pi$ (more specifically, its parameters like location and scale parameters in Gaussian) is directly calculated for the gradient-based optimization.

Finally, how to automatically tune $\alpha$, given as a constant parameter so far, is shown since its fine-tuning for each task is critical but time consuming.
This is a new definition in the literature written by the same authors of SAC~\cite{haarnoja2018soft2}.
Specifically, the optimization problem of SAC is first reinterpreted as a constrained optimization problem with respect to the policy entropy.
\begin{align}
    \pi^{\ast} &= \arg\min_{\pi} \mathbb{E}_{s \sim \mathcal{D}, \rho^\pi}
    \left[ - \sum_{k=0}^{\infty} \gamma^{k} r(s_{t+k}, a_{t+k}) \mid s \right ]
    \\
    \mathrm{s.t.} \ & \
    \mathcal{H}(\pi(\cdot \mid s)) \geq \mathcal{H}^{\ast} \ ({}^\forall s \in \mathcal{D})
    \nonumber
\end{align}
where $\mathcal{H}^{\ast}$ denotes the lower bound, which is often designed as $- |\mathcal{A}|$ for the continuous action space.

In that case, $\alpha$ is given as the Lagrange multiplier when reverting to the usual SAC optimization problem, and relying on this new interpretation, $\alpha$ can be optimized in the following minimization problem.
\begin{align}
    \alpha^{\ast} = \arg\min_{\alpha}\mathbb{E}_{s \sim \mathcal{D}, a \sim \pi}[ - \alpha \{ \ln \pi(a \mid s) + \mathcal{H}^{\ast} \}]
    \label{eq:auto_tune_conv}
\end{align}
Here, $\pi$ in this problem should be $\pi^{\ast}$ in theory, but since $\pi^{\ast}$ is not given explicitly, it is substituted by $\pi$ in practice.
As a result, qualitatively, if $\ln\pi > \mathcal{H}^{\ast}$, $\alpha$ will be increased to maximize entropy;
otherwise, if $\ln\pi < \mathcal{H}^{\ast}$, $\alpha$ will be decreased to allow smaller entropy.

Note that, in order to satisfy $\alpha \geq 0$, $\tilde{\alpha} \in \mathbb{R}$ (e.g. $\alpha = e^{\tilde{\alpha}}$) is prepared as a trainable parameter and updated by using the mirror descent method~\cite{beck2003mirror}.
That is, $\tilde{\alpha}$ is updated using the gradient of eq.~\eqref{eq:auto_tune_conv} with respect to $\alpha$, $\mathbb{E}_{s \sim \mathcal{D}, a \sim \pi}[ - \{ \ln \pi(a \mid s) + \mathcal{H}^{\ast} \}]$, and then transformed again to $\alpha$.
This avoids the gradient vanishing problem in $\alpha \simeq 0$ due to the nonlinear function to satisfy $\alpha \geq 0$.

%%%%%%%%%%%%%%%%%%%%%%%%%%%%%%%%%%%%%%%%%%%%%%%%%%%%%%%%%%%%%%%%%%%%%%%%%%%%%%%%
\section{Proposal}

%%%%%%%%%%%%%%%%%%%%%%%%%%%%%%%%%%%%%%%%%%%%%%%%%%%%%%%
\subsection{Open issue in SAC}

It has been reported that SAC acquires higher control performance (e.g. higher profitability and robustness) than previous RL algorithms.
This is said to be mainly due to maximizing the policy entropy $\mathcal{H}(\pi)$ as much as possible without losing the orignal purpose (i.e. reward).
However, the actual implementation introduced in the above does not necessarily provide the maximal $\mathcal{H}(\pi)$.

This is because the auto-tuned temperature parameter $\alpha$, as defined in eq.~\eqref{eq:auto_tune_conv}, only makes $\mathcal{H}(\pi)$ (or $- \ln \pi$) closer to $\mathcal{H}^{\ast}$, as expected from its qualitative behavior (also reported in the literature~\cite{xu2021target}).
In other words, the above auto-tuning is regarded to be derived based on the method of Lagrange multiplier with \textit{equality} constraint, which is inconsistent with the maximization of $\mathcal{H}(\pi)$.
Therefore, even though the optimization problem was originally designed to maximize $\mathcal{H}(\pi)$ with \textit{inequality} constraint, the implementation was for $\mathcal{H}(\pi) = \mathcal{H}^{\ast}$.
Because of this behavior, $\mathcal{H}^{\ast}$ is often called ``target'' entropy, although it should be defined as the lower bound.
The limitation of $\alpha \geq 0$ may cause $\mathcal{H}(\pi) > \mathcal{H}^{\ast}$ without fully satisfying this equality constraint, but in practice, the current auto-tuning makes $\mathcal{H}(\pi)$ mostly converge to $\mathcal{H}^{\ast}$ in the training (see the simulation results later).
As a result, without an appropriate $\mathcal{H}^{\ast}$, the policy entropy is expected to be too small to achieve the expected performance, and the full fruits of maximizing the policy entropy to be obtained may not be appreciated.

%%%%%%%%%%%%%%%%%%%%%%%%%%%%%%%%%%%%%%%%%%%%%%%%%%%%%%%
\subsection{Auto-tuned temperature with slack variable}

To resolve this issue, this study transforms the troublesome inequality constraint into the simple equality constraint in another way.
This allows the method of Lagrange multiplier to be used with the transformed equality constraint as in the conventional auto-tuning of $\alpha$.
For this purpose, a slack variable $\Delta \geq 0$ is commonly introduced~\cite{boyd2004convex} (even for the recent machine learning~\cite{picheny2016bayesian}).
\begin{align}
    \mathcal{H}(\pi(\cdot \mid s)) &\geq \mathcal{H}^{\ast} \quad ({}^\forall s \in \mathcal{D})
    \nonumber \\
    \Leftrightarrow
    \mathcal{H}(\pi(\cdot \mid s)) &= \mathcal{H}^{\ast} + \Delta(s) \quad ({}^\forall s \in \mathcal{D}, \ \Delta \geq 0)
    \label{eq:def_slack}
\end{align}
In short, by finding $\Delta$ that holds $\mathcal{H}(\pi) - \mathcal{H}^{\ast} = \Delta$ at any states, the inequality constraint can be theoretically transformed into the equality constraint.

Given the appropriate $\Delta$, the optimization of $\alpha$ can be renewed from eq.~\eqref{eq:auto_tune_conv} to the following one.
\begin{align}
    \alpha^{\ast} = \arg\min_{\alpha}\mathbb{E}_{s \sim \mathcal{D}, a \sim \pi}[ - \alpha \{ \ln \pi(a \mid s) + \mathcal{H}^{\ast} + \Delta(s) \}]
    \label{eq:loss_alpha}
\end{align}
As a simple interpretation of this update rule, $\alpha$ makes $\mathcal{H}(\pi)$ (or $-\ln\pi$) converge to $\mathcal{H}^{\ast} + \Delta \geq \mathcal{H}^{\ast}$, which is newly given as a target value (previously, $\mathcal{H}^{\ast}$ was that role, not the lower bound of the policy entropy).
In addition, since $\Delta$ is regarded as a state-dependent function, the different policy entropy for each state can be expected:
for example, it may be small for scenes requiring accuracy and; large for scenes requiring exploration.

%%%%%%%%%%%%%%%%%%%%%%%%%%%%%%%%%%%%%%%%%%%%%%%%%%%%%%%
\subsection{Design of slack variable}

%Figure
\begin{figure}[tb]
    \centering
    \includegraphics[keepaspectratio=true,width=0.96\linewidth]{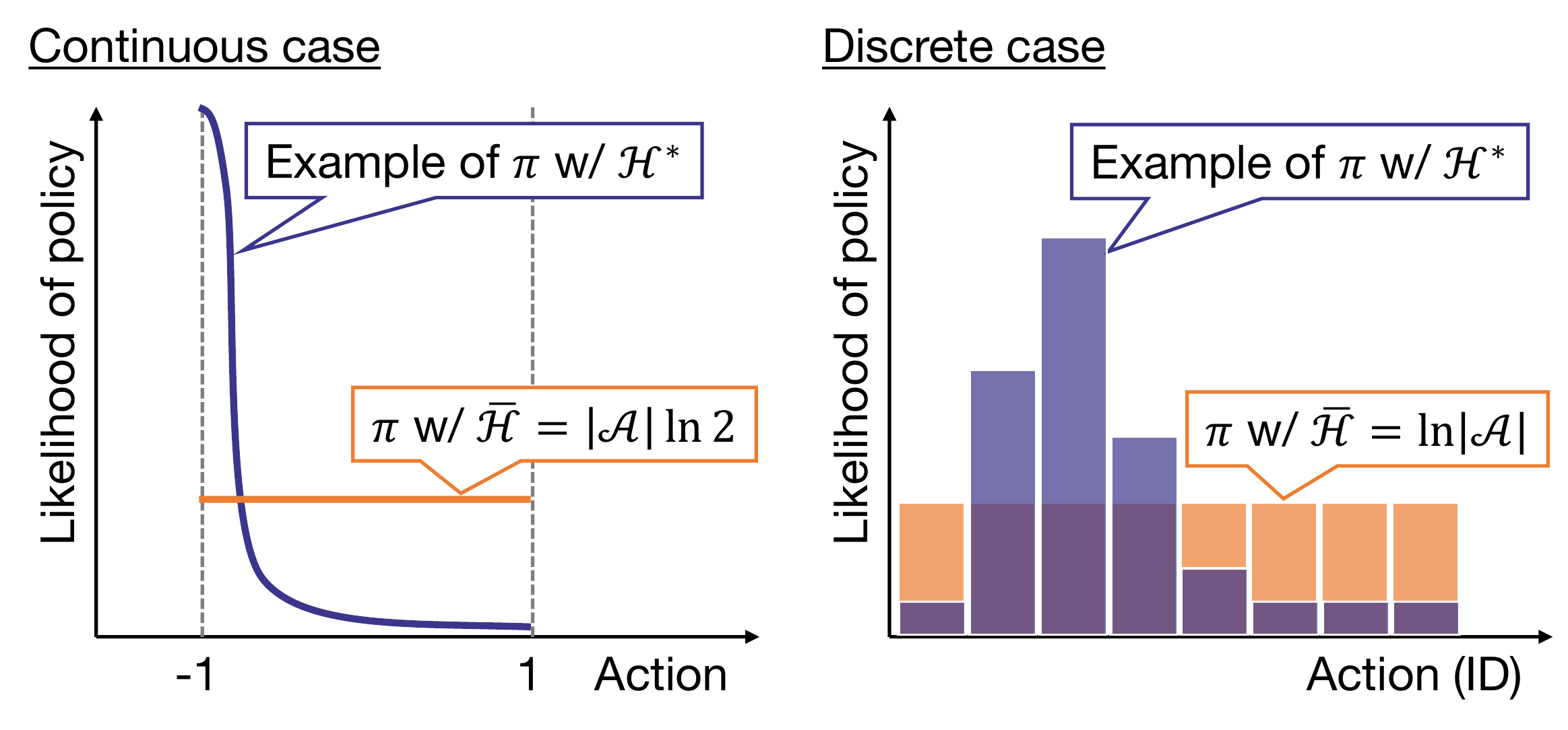}
    \caption{Upper bounds for policy entropy}
    \label{fig:entropy_upper_bound}
\end{figure}

$\Delta$ is specifically designed as a state-dependent function.
It is approximated by deep neural networks with state input.
Since the output from deep neural networks, $d$, is real-valued, a mapping to the domain of $\Delta$ is required.
In the general slack variable, $\Delta$ is defined to be non-negative.
In addition, in the case of SAC, its upper bound is also found.

Specifically, SAC mainly targets a bounded continuous action space, which is popular in many tasks, and explicitly transforms the action space so that $\mathcal{A} = [-1, 1]^{|\mathcal{A}|}$.
In such a box-constrained space, the policy can be uniformly distributed, $\pi = 2^{-|\mathcal{A}|}$, with clearly maximal entropy (i.e. $\overline{\mathcal{H}} = |\mathcal{A}| \ln 2$ illustrated in the left of Fig.~\ref{fig:entropy_upper_bound}).
Similarly, even in a discrete action space, the uniform distribution, $\pi = |\mathcal{A}|^{-1}$, maximizes entropy (i.e. $\overline{\mathcal{H}} = \ln |\mathcal{A}|$ illustrated in the right of Fig.~\ref{fig:entropy_upper_bound}).

In summary, the upper bound of $\Delta$ for the policy entropy in SAC, $\overline{\Delta}$, is obtained as follows:
\begin{align}
    \overline{\Delta} =
    \begin{cases}
        |\mathcal{A}| \ln 2 - \mathcal{H}^{\ast} & \mathrm{(Continuous)}
        \\
        \ln |\mathcal{A}| - \mathcal{H}^{\ast} & \mathrm{(Discrete)}
    \end{cases}
\end{align}
where $\mathcal{H}^{\ast}$ for the continuous action space is recommended to be $- |\mathcal{A}|$ in the original paper~\cite{haarnoja2018soft2} and $\mathcal{H}^{\ast}$ for the discrete one is tuned as $0.98 \ln |\mathcal{A}|$ in SAC-Discrete~\cite{christodoulou2019soft}, although they are for the conventional equality constraint.
Please note that $\mathcal{H}^{\ast}$ has to be given such that $\overline{\Delta} \geq 0$.
Although the recommended values so far satisfy that condition, it would be desired to find new recommendations since the previous $\mathcal{H}^{\ast}$ did not play the role of the lower bound.

With $\overline{\Delta}$ and the fact that $\Delta \geq 0$, the real-valued output from the corresponding networks, $d$, can be mapped to $\Delta$ within its domain properly by a sigmoid function and multiplying with $\overline{\Delta}$.
That is, $d$ is transformed to $\Delta \in [0, \overline{\Delta}]$ by the following equation.
\begin{align}
    \Delta = \overline{\Delta} \sigma(d)
\end{align}
where $\sigma(\cdot)$ denotes the sigmoid function.
Note that although $\sigma(d) = 1 / (1 + \exp(-d))$ is the most representative design, this work employs the first derivative of the squareplus function from the literature~\cite{barron2021squareplus,kobayashi2023design}, which has heavy tails with higher numerical resolution in the neighborhood of the boundaries (i.e. $0$ and $\overline{\Delta}$).

%%%%%%%%%%%%%%%%%%%%%%%%%%%%%%%%%%%%%%%%%%%%%%%%%%%%%%%
\subsection{Optimization of slack variable}

The designed $\Delta \in [0, \overline{\Delta}]$ is optimized through minimization of an appropriate loss function.
In general, Lagrangian function (a.k.a. the loss function for $\pi$ in eq.~\eqref{eq:loss_pi}) should also be minimized with regard to $\Delta$.
In other words, $\alpha \Delta$ should be minimized for eliminating $\Delta$, the speed of which depends on $\alpha$, and for checking the performance on the boundary (i.e. the lower bound).
In particular, when $\alpha$ is small enough to allow $\mathcal{H}(\pi)$ to decrease, the decrease of $\Delta$ will be slowed down to moderate the decrease of $\mathcal{H}(\pi)$.

Another requirement is to minimize the error between the left and right sides of eq.~\eqref{eq:def_slack} to satisfy the transformed equality constraint.
This can be done simply by minimizing the absolute error between the left and right sides of eq.~\eqref{eq:def_slack}.
However, if the equality constraint is sufficiently satisfied only by the optimization of $\Delta$, there will no longer be room for the auto-tuning of $\alpha$.
To avoid this issue, $\epsilon$-insensitive loss~\cite{vapnik1999nature} is employed.
That is, if the error is below a certain level $\epsilon \geq 0$, the optimization of $\Delta$ for satisfying the equality constraint is stopped.

Finally, it is necessary to set up a balance between the above two loss functions.
The simple solution would be the weighted sum, but this would require the introduction and adjustment of a new weight hyperparameter.
Instead of the weighted sum, a mechanism to switch between the two loss functions depending on $\epsilon$ is considered.
That is, when the equality constraint is not achieved up to a certain level of $\epsilon$, $\Delta$ is optimized only for satisfying the equality constraint, otherwise, $\Delta$ is gradually reduced to zero.

\begin{figure}[tb]
    \centering
    \includegraphics[keepaspectratio=true,width=0.96\linewidth]{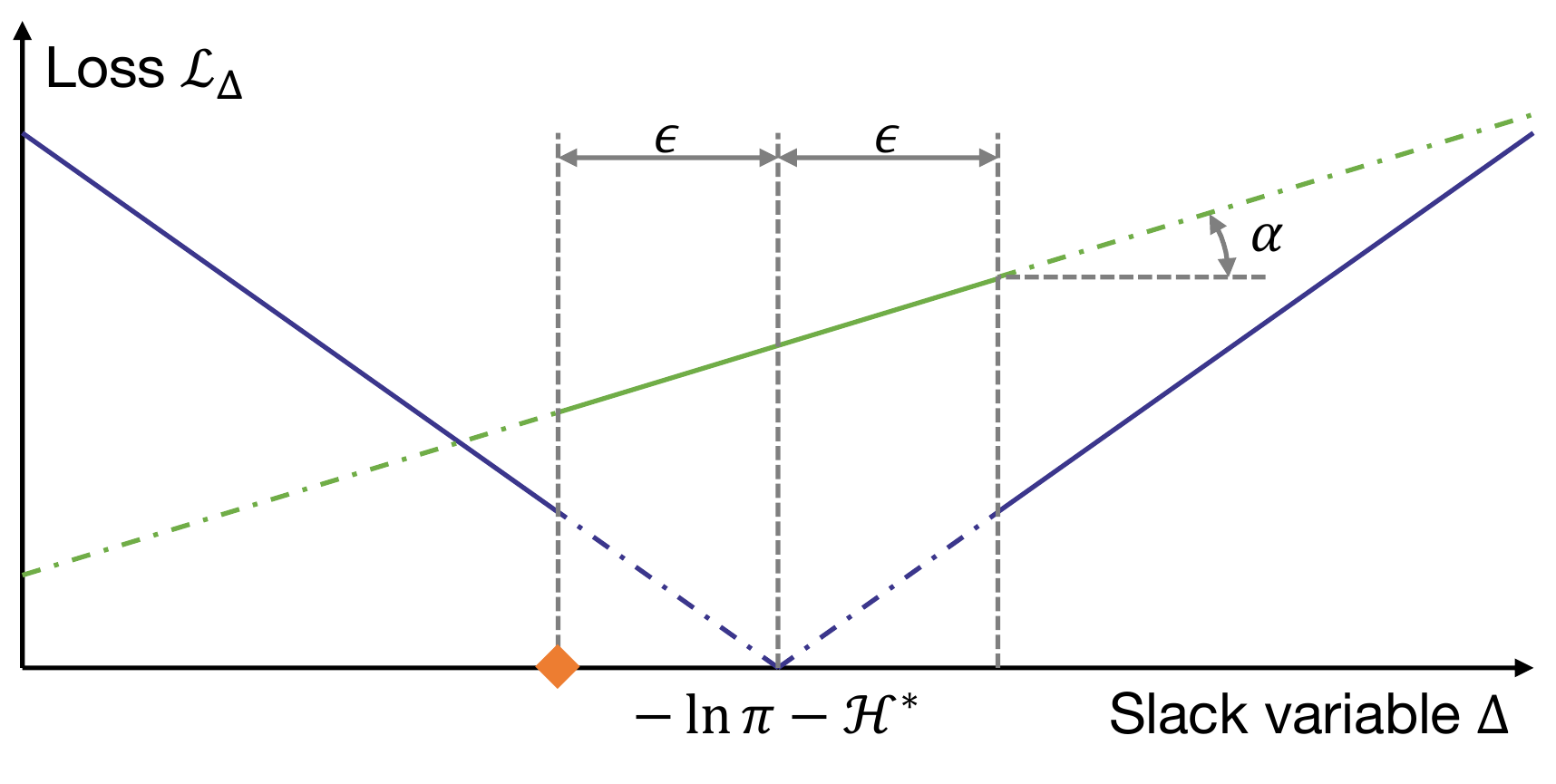}
    \caption{Loss function for slack variable $\Delta$}
    \label{fig:loss_slack}
\end{figure}

In summary, $\Delta$ is optimized through the following minimization problem.
\begin{align}
    \Delta^{\ast} &= \arg\min_{\Delta} \mathbb{E}_{s \sim \mathcal{D}, a \sim \pi} [\mathcal{L}_{\Delta}(s, a; \epsilon)]
    \label{eq:loss_delta} \\
    \mathcal{L}_{\Delta}(s, a; \epsilon) &=
    \begin{cases}
        |\ln \pi(a \mid s) + \mathcal{H}^{\ast} + \Delta(s)| & |e| > \epsilon
        \\
        \alpha \Delta(s) & |e| \leq \epsilon
    \end{cases}
    \nonumber
\end{align}
where $|e| = |\ln \pi(a \mid s) + \mathcal{H}^{\ast} + \Delta(s)|$.
By minimizing this loss function, $\ln \pi + \mathcal{H}^{\ast} + \Delta = - \epsilon$ will be satisfied if $\ln \pi$ is stationary, as shown in Fig.~\ref{fig:loss_slack}.
In practice, however, the equilibrium point is shifted by $\ln\pi$ according to the automatically tuned $\alpha$, and $\Delta$ is optimized accordingly, which is an aspect of alternating optimization~\cite{bezdek2003convergence}.

As a remark, in the implementation of this paper, the above minimization problem is solved by using the mirror descent method following the optimization of $\alpha$.
That is, the gradient of $\mathcal{L}_{\Delta}$ w.r.t. $\Delta$ is utilized to update $d$ (i.e. the value before mapping to $\Delta$) and network parameters involved in $d$.
This can be accomplished by redefining eq.~\eqref{eq:loss_delta} as the following minimization problem.
\begin{align}
    d^{\ast} &= \arg\min_{d} \mathbb{E}_{s \sim \mathcal{D}, a \sim \pi} [\mathcal{L}_{d}(s, a; \epsilon)]
    \label{eq:loss_delta_mirror} \\
    \mathcal{L}_{d}(s, a; \epsilon) &=
    \begin{cases}
        \mathrm{sign}(\ln \pi(a \mid s) + \mathcal{H}^{\ast} + \Delta(s)) d(s) & |e| > \epsilon
        \\
        \alpha d(s) & |e| \leq \epsilon
    \end{cases}
    \nonumber
\end{align}
where $\mathrm{sign}(\cdot)$ denotes the sign function.
This implementation would alleviate the gradient vanishing problem due to $\sigma$, which maps $d$ to $\Delta$.

%Figure

%%%%%%%%%%%%%%%%%%%%%%%%%%%%%%%%%%%%%%%%%%%%%%%%%%%%%%%%%%%%%%%%%%%%%%%%%%%%%%%%
\section{Simulations}

%%%%%%%%%%%%%%%%%%%%%%%%%%%%%%%%%%%%%%%%%%%%%%%%%%%%%%%
\subsection{Simulation conditions}

To statistically verify the proposed SAC improvements, dynamical simulations are performed on Mujoco~\cite{todorov2012mujoco} and Pybullet~\cite{coumans2016pybullet}.
The three types of tasks implemented on both simulators are conducted:
\begin{itemize}
    \item \textit{Hopper} (Hopper-v4/HopperBulletEnv-v0)
    \item \textit{HalfCheetah} (HalfCheetah-v4/HalfCheetahBulletEnv-v0)
    \item \textit{Ant} (Ant-v4/AntBulletEnv-v0)
\end{itemize}
where the first and second tags in each parenthesis denotes the tasks implemented in Mujoco and Pybullet, respectively.
To easily distinguish between the two simulators, the task names end with ``M'' for Mujoco ``B'' for Pybullet.
Each task is trained for 2000 episodes.

The following three conditions are compared:
\begin{itemize}
    \item Conventional case: $\mathcal{H}^{\ast} = - |\mathcal{A}|$ with $\Delta = 0$ (i.e. $\mathcal{H}(\pi) \in [-|\mathcal{A}|, -|\mathcal{A}|]$)
    \item Proposed case 1: $\mathcal{H}^{\ast} = - |\mathcal{A}|$ with $\Delta \in [0, \overline{\Delta}]$ (i.e. $\mathcal{H}(\pi) \in [-|\mathcal{A}|, |\mathcal{A}| \ln 2]$)
    \item Proposed case 2: $\mathcal{H}^{\ast} = \overline{\mathcal{H}} - 2|\mathcal{A}|$ with $\Delta \in [0, \overline{\Delta}]$ (i.e. $\mathcal{H}(\pi) \in [|\mathcal{A}|(\ln 2 - 2), |\mathcal{A}| \ln 2]$)
\end{itemize}
The first corresponds to the conventional method, the second is the proposed method using the same lower bound as the conventional method~\cite{haarnoja2018soft2}, and the third employs a lower bound smaller than the conventional recommended value (i.e. $\mathcal{H}^{\ast} \simeq -1.3 |\mathcal{A}|$).
By comparing the first and the others, it can be confirmed that the inequality constraint is properly satisfied by the proposed slack variable.
By comparing the second and the third, the performance difference due to the specified lower bound is qualitatively analyzed.
Note that all other configurations are the same, as described in Appendix~\ref{app:impl}.

A total of 24 trials are trained with different random seeds for each combination of the above tasks and methods.
In order to statistically evaluate the performance of the acquired policy, each learned policy is tested 100 times on the corresponding task with adversarial attacks inspired by the literature~\cite{tessler2019action}.
Specifically, the actions are replaced by ones generated from Gaussian noise and transformed within $[-0.2, 0.2]$ with a probability of 5~\% for Hopper and 20~\% for HalfCheetah and Ant.
The following two criteria of robustness are considered:
one is the averaged norm of action, which suggests whether the acquired policy behaves conservatively;
and another is the sum of rewards during each test, which shows whether the acquired policy maintains the control performance.

%Figure
\begin{figure*}[tb]
    \centering
    \includegraphics[keepaspectratio=true,width=0.96\linewidth]{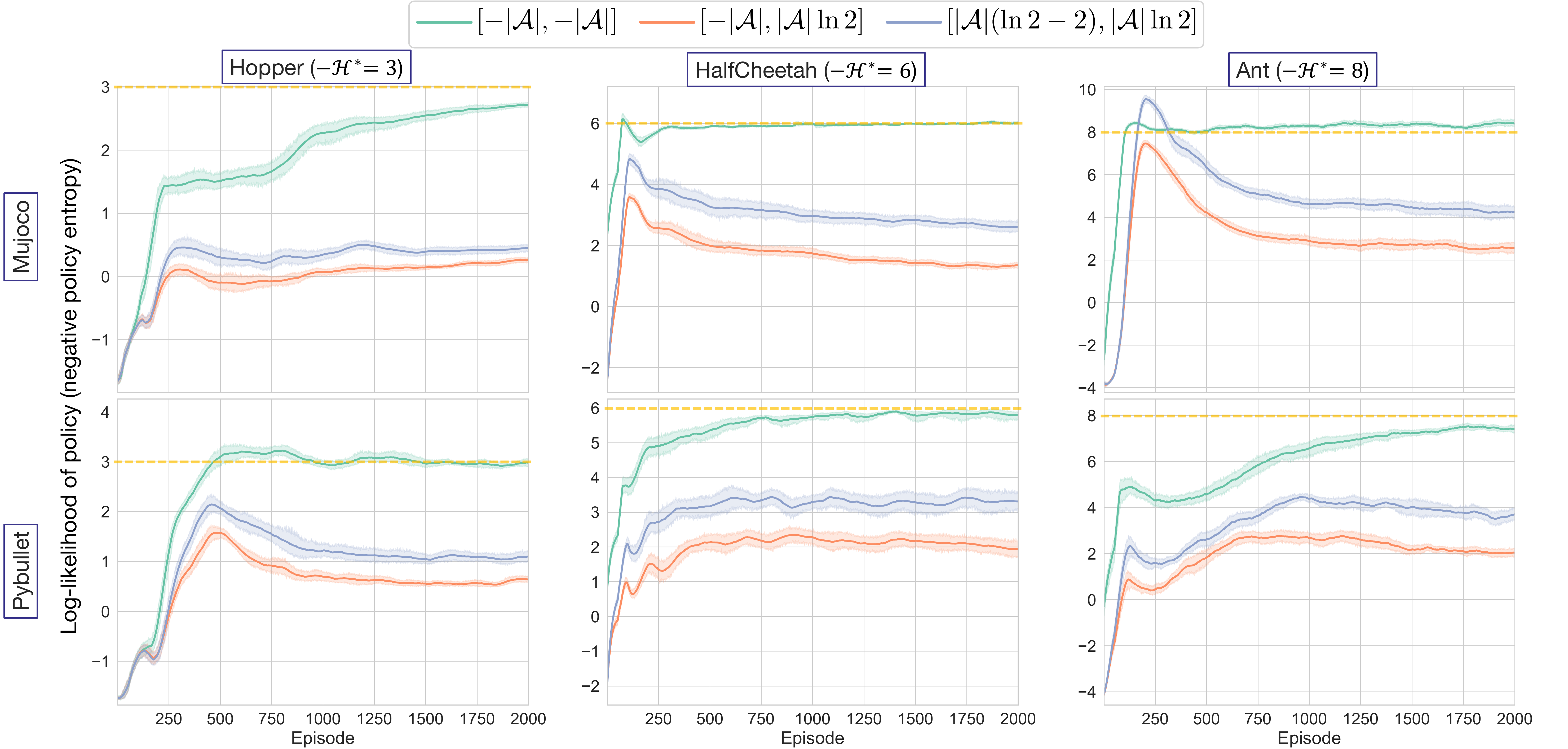}
    \caption{Learning curves of the log-likelihood of the policy $\ln \pi \simeq - \mathcal{H}(\pi)$}
    \label{fig:learn_logpi}
\end{figure*}

%Figure
\begin{figure*}[tb]
    \centering
    \includegraphics[keepaspectratio=true,width=0.96\linewidth]{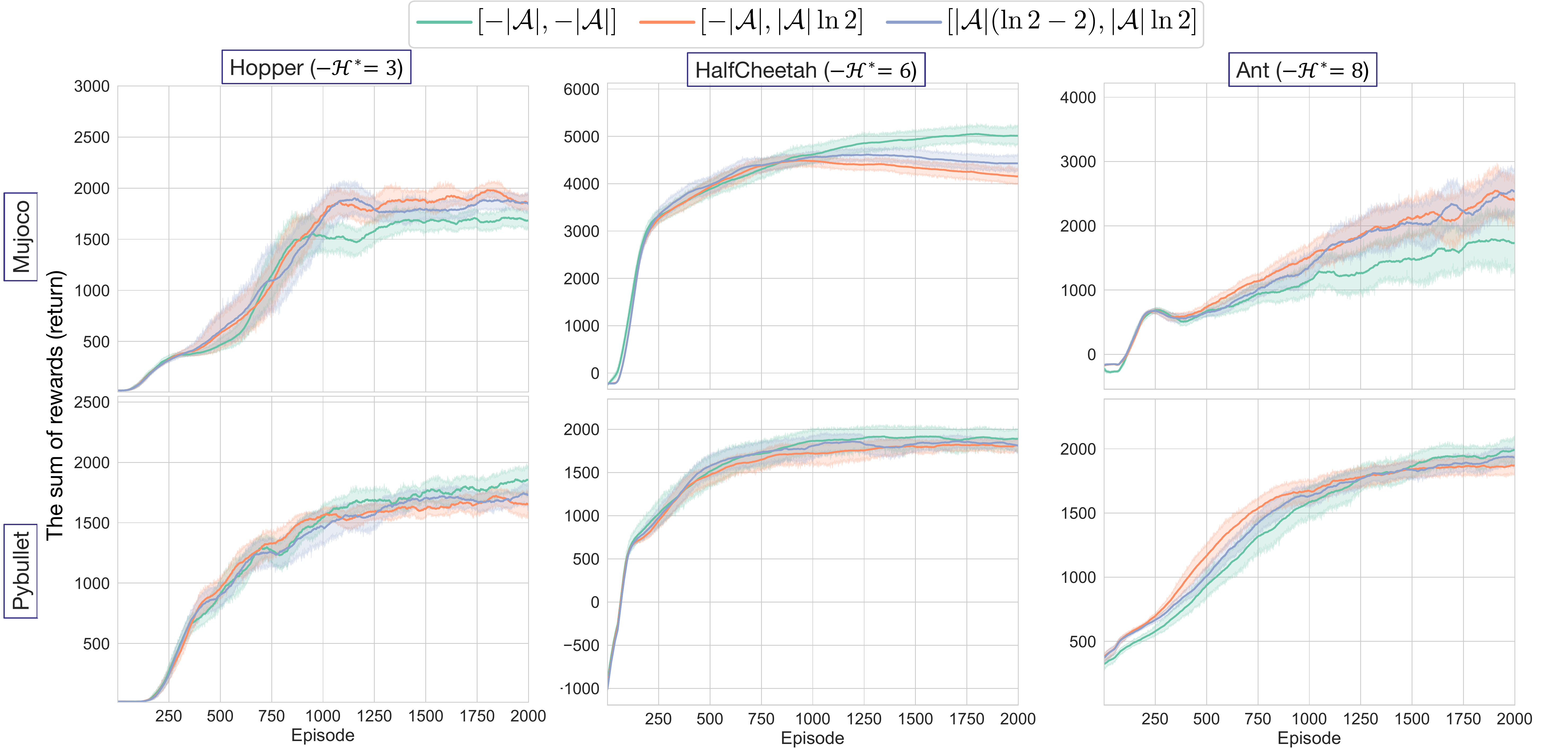}
    \caption{Learning curves of the return $R$}
    \label{fig:learn_score}
\end{figure*}

%%%%%%%%%%%%%%%%%%%%%%%%%%%%%%%%%%%%%%%%%%%%%%%%%%%%%%%
\subsection{Learning tendency}

The episodic average of log-likelihood of the policy $\ln \pi$ (corresponding to the negative policy entropy $-\mathcal{H}(\pi)$) is shown in Fig.~\ref{fig:learn_logpi}.
In addition, the learning curves of the return $R$ are depicted in Fig.~\ref{fig:learn_score}.
Note that they were obtained using the sampled actions from the optimizing policy $\pi$.

First of all, the results show that $\ln \pi$ in the conventional SAC was approximately stuck to $|\mathcal{A}|$, and that the optimization of $\alpha$ satisfied the equality constraint for the lower bound $\mathcal{H}^{\ast} = - |\mathcal{A}|$.
On the other hand, the proposed method clearly reduced $\ln \pi$ on average while allowing for fluctuation more, as can be found fron the width of the respective confidence intervals.
Namely, the proposed method made room for increase/decrease in the policy entropy according to situations.
The proposed method not only achieved such adaptability of the policy entropy due to holding the inequality constraint, but kept the policy entropy above the lower bound (on average).
Note that the fact that the third condition has a smaller policy entropy than the second one in generally parallel to the second one could be due to some kind of initial value dependence, although it is also possible that the smaller lower bound has simply lowered the total.
That is, if deep neural networks outputting $\Delta$ are initialized with the same random seed, the initial difference in output is proportional to $\overline{\Delta}$, so the initial point of search for optimizing $\Delta$ may have changed and found different equilibrium points.

The benefit of the adaptive policy entropy was confirmed in HopperM and AntM, where $R$ was improved from the conventional SAC by suppressing over-learning.
In addition, more efficient exploration can be found in HalfCheetahM and AntB in the early stages of learning.
On the other hand, it was found that higher policy entropy sometimes inhibited the performance improvements due to less deterministic actions.
$R$, especially in HalfCheetahM, actually decreased as the policy entropy increased in the proposed method, while the third condition, in which the lower bound is lowered, mitigated this adverse effect by preventing the policy entropy from becoming excessively large.
Nevertheless, the verification of the boundary (i.e. the lower bound), which is one of the objectives imposed on the optimization of $\Delta$ in the proposed method, has not been fully achieved.
A more sophisticated design of the loss function would be necessary in the future.

%%%%%%%%%%%%%%%%%%%%%%%%%%%%%%%%%%%%%%%%%%%%%%%%%%%%%%%
\subsection{Test results}

%Figure
\begin{figure}[tb]
    \centering
    \includegraphics[keepaspectratio=true,width=0.96\linewidth]{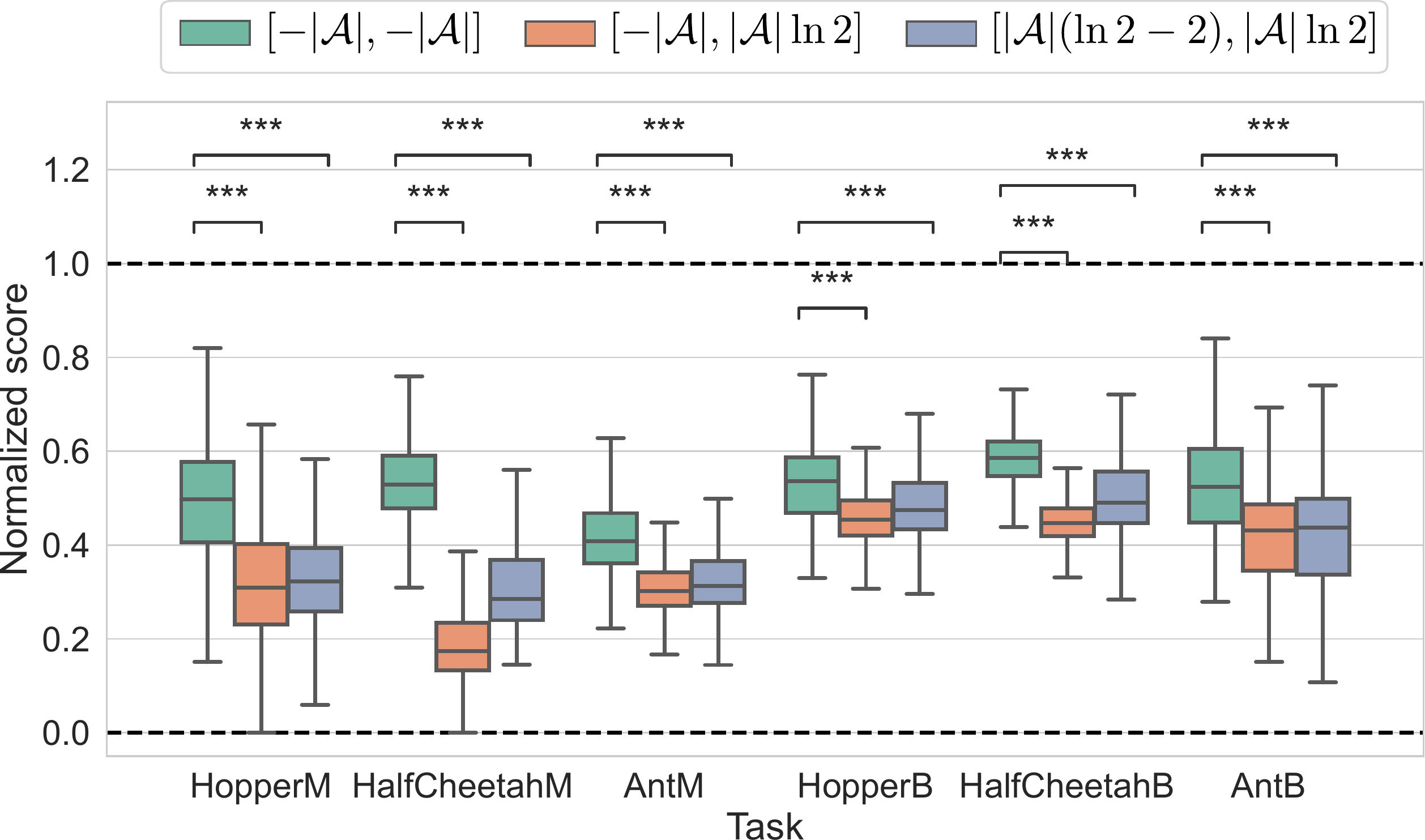}
    \caption{Normalized norm of action on the perturbed tests}
    \label{fig:robust_action_stat}
\end{figure}

%Figure
\begin{figure}[tb]
    \centering
    \includegraphics[keepaspectratio=true,width=0.96\linewidth]{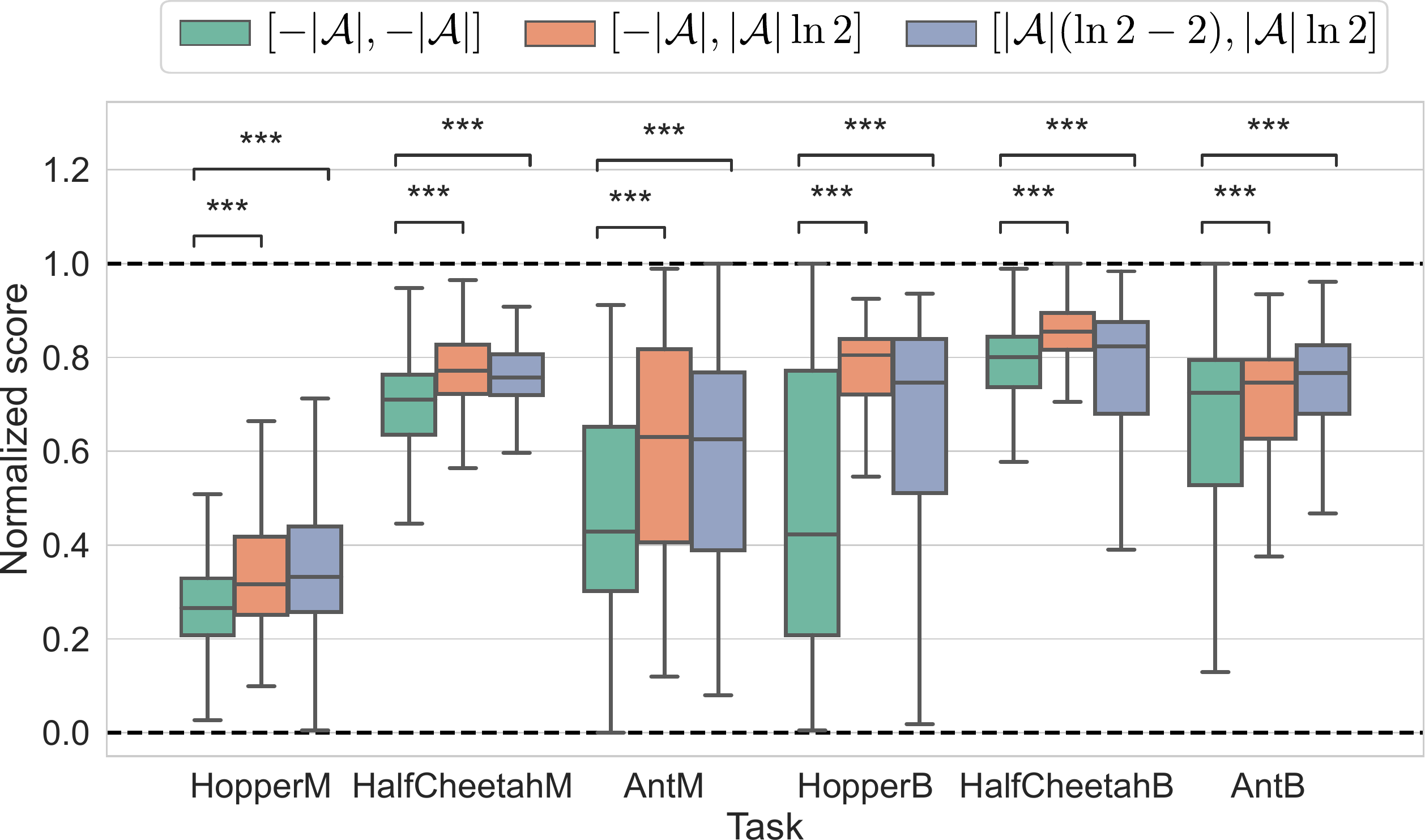}
    \caption{Normalized return on the perturbed tests}
    \label{fig:robust_result_stat}
\end{figure}

The test results with adversarial attacks are investigated to evaluate the statistical performance (especially robustness) of the acquired policy by each condition.
First, stochastic models that are robust to noise and perturbation generally tend to be conservative (e.g. their output is close to zero)~\cite{ben2009robust}.
This can be explained more directly when the entropy maximization is applied to generate bounded stochastic values, as in SAC.
That is, since the likelihood of actions near the boundary is necessarily high, attempting to suppress them by maximizing the policy entropy forces the actions to move away from the boundary and thus toward the center (i.e. zero).
From the viewpoint of robot control, this conservativeness implicitly improves energy efficiency and smoothes out behaviors, as like the literature~\cite{kobayashi2022l2c2}.

Here, the episodic average of the L2 norm of action is depicted in Fig.~\ref{fig:robust_action_stat}.
Note that statistical significance is confirmed using \texttt{mannwhitneyu(alternative="less")} in \texttt{scipy.stats} package, which confirms whether the first condition is greater than the others.
The results are clear at a glance:
the norm of action is inversely related to the policy entropy.
In other words, the two methods that succeeded in increasing the policy entropy compared to the conventional method did indeed achieve conservative behaviors.
However, they are not simply inversely proportional.
For example, in AntM and AntB, the difference in the norm of action was small, even though the policy entropy was somewhat distant between the second and third conditions.
This fact is expected because the optimal behaviors in AntM and AntB would undergo a phase transition, as the policy entropy can now be larger than the lower bound, correctly reflecting the inequality constraint.
The conventional method would not only lose the conservative behavior by the maximized policy entropy, but also overlook such an optimal solution.

The acquired policy is not said to be robust if it is only conservative in its behaviors.
It must be able to recover when the agent deviates from the optimal trajectory due to adversarial attacks.
This ability can be confirmed by evaluating the control performance (i.e. the return $R$) under adversarial attacks.
Since the agent with the entropy-maximized policy should encounter various situations and learn how to recover from them during training with sampled actions.
That is, the entropy-maximized policy is expected to be robust, as suggested in the literature~\cite{haarnoja2018soft2,eysenbach2021maximum}.

Here, the return is statistically depicted in Fig.~\ref{fig:robust_result_stat}.
Note that statistical significance is confirmed using \texttt{mannwhitneyu(alternative="greater")}, which confirms whether the first condition is less than the others.
As a result, the conventional method, which was sometimes superior to the others as shown in Fig.~\ref{fig:learn_score}, was vulnerable to the adversarial attacks and statistically performed worse than the proposed method.
The larger the policy entropy, the more robust the control performance would be, as suggested by the fact that the control performance was generally maintained.
However, the third condition, with a smaller policy entropy, had a higher original control performance without the adversarial attacks than the second condition, and the median, Q1, and Q3 were higher for HopperM and AntB.
Therefore, the third condition with a smaller lower bound, which provides a better balance of the policy entropy, would be more preferable, especially if the verification function of the lower bound is well performed with some refinement.

%%%%%%%%%%%%%%%%%%%%%%%%%%%%%%%%%%%%%%%%%%%%%%%%%%%%%%%%%%%%%%%%%%%%%%%%%%%%%%%%
\subsection{Demonstrations with real robot}

%%%%%%%%%%%%%%%%%%%%%%%%%%%%%%%%%%%%%%%%%%%%%%%%%%%%%%%
\subsection{Setup}

%Figure
\begin{figure}[tb]
    \centering
    \includegraphics[keepaspectratio=true,width=0.64\linewidth]{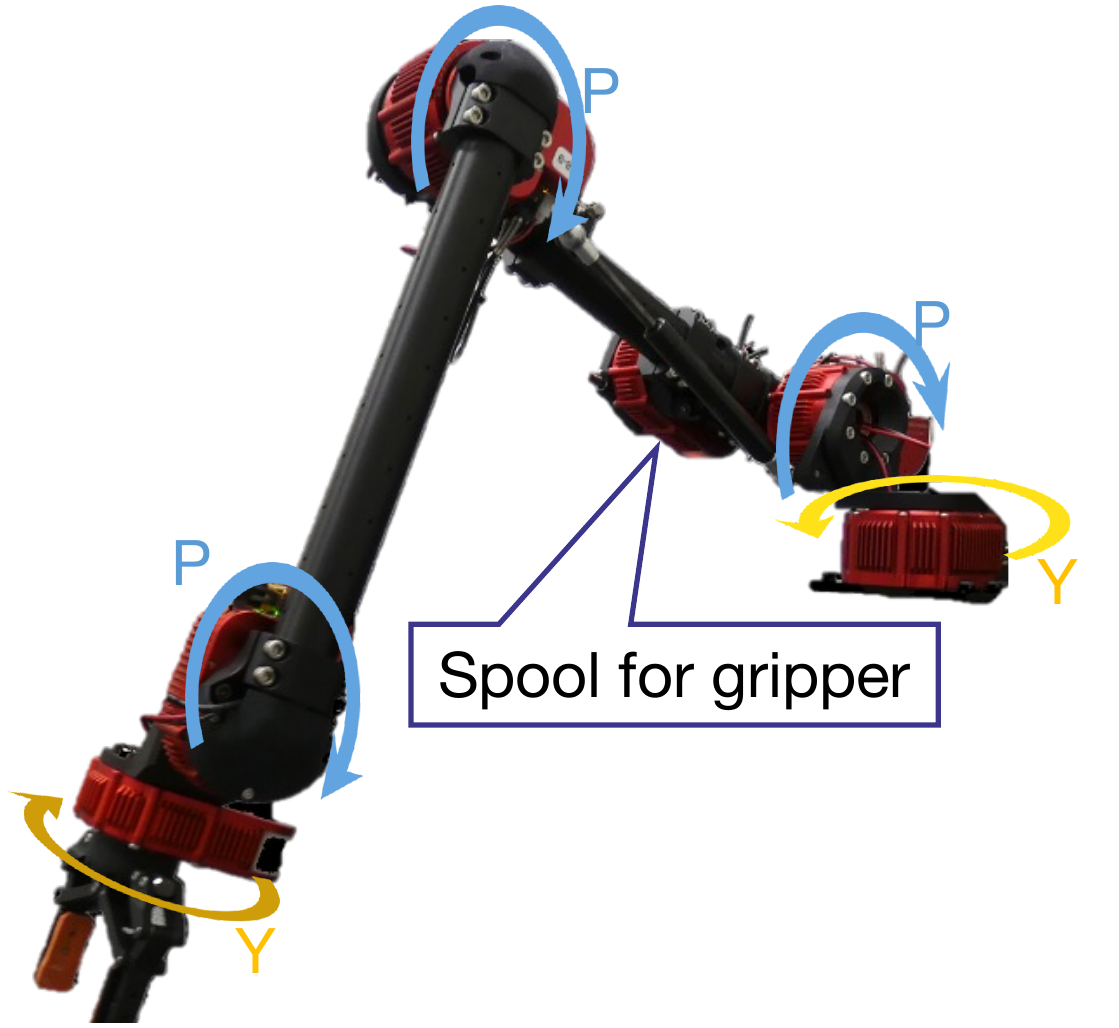}
    \caption{Manipulator used in the variable impedance task}
    \label{fig:hebi_arm}
\end{figure}

%Figure
\begin{figure}[tb]
    \centering
    \includegraphics[keepaspectratio=true,width=0.96\linewidth]{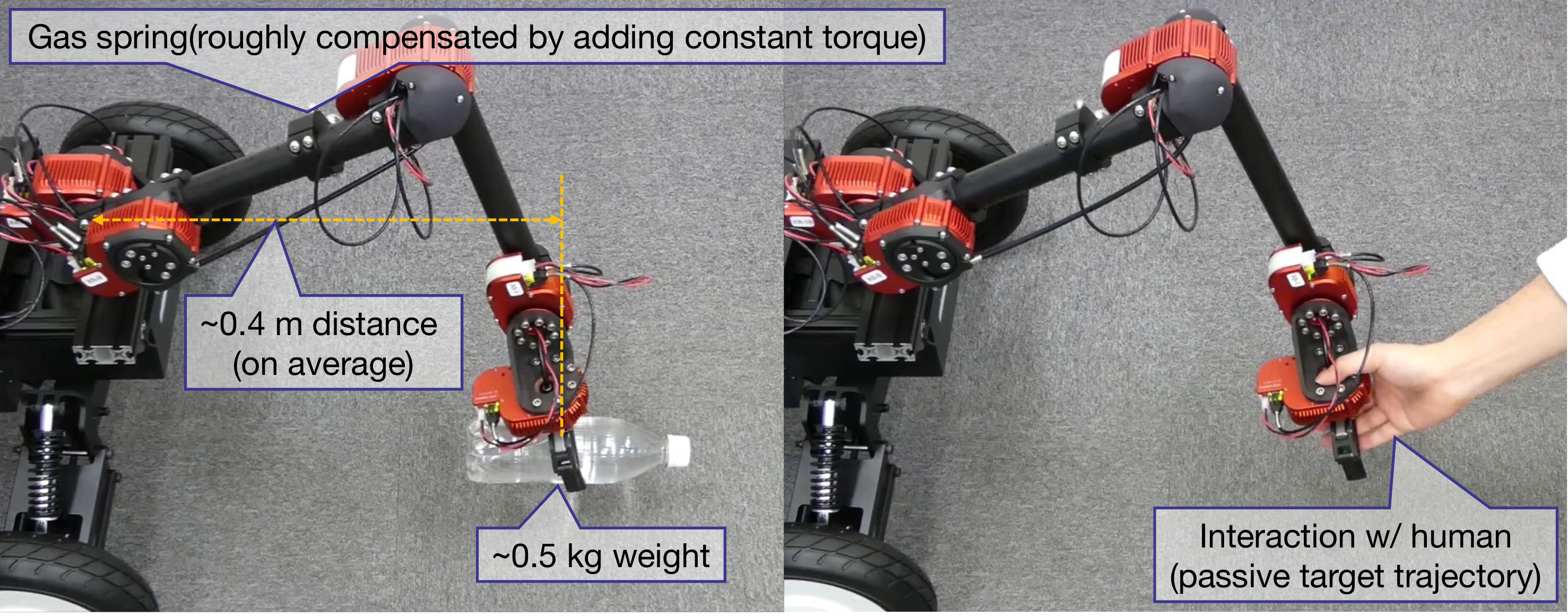}
    \caption{\textit{Weight} and \textit{Interaction} scenarios}
    \label{fig:hebi_test}
\end{figure}

Through the above simulations, it can be confirmed that the proposed method is more robust to pertubation than the conventional SAC by truly maximizing the policy entropy.
For the next step, the practical usefulness of the proposed method in real-world robot control is demonstrated.
The robot used in this experiment is a 5-DoF manipulator as shown in Fig.~\ref{fig:hebi_arm}.
When tracking a randomly-given trajectory for the robot's end-effector, six impedance parameters are optimized using RL as a variable impedance control system~\cite{buchli2011learning,ficuciello2015variable}.
This optimization aims to minimize efforts while keeping the trajectory error within a threshold.
The details of the experimental configuration are summarized in Appendix~\ref{app:robot}.

During/after training five models with different random seeds for each method, each episode is evaluated with the following criteria:
i.e. return; tracking error; frequency of threshold exceeded; exerted effort; norm of action; and log-likelihood of policy.
The first four are criteria for the control objectives, and the remaining two are criteria for the robustness of the policy.
To show the robustness of the policy trained by the proposed method, three scenarios (see Fig.~\ref{fig:hebi_test}) are tested with 10 trials for each.
\begin{itemize}
    \item \textit{Vanilla}: A random trajectory, which was not contained in the training phase, is tracked.
    \item \textit{Weight}: The end-effector grasps a roughly 0.5~kg weight, while adjusting an offset torque on a shoulder.
    \item \textit{Interaction}: The target position of the end-effector is given passively while a human moves the end-effector through physical interaction, while its range is limited to the experienced range during training.
\end{itemize}
At \textit{Weight} scenario, the offset torque (see Appendix~\ref{app:robot}) is adjusted to roughly compensate the weight, which causes roughly $0.5$~kg $\times -9.8$~m/s$^2$ $\times 0.4$~m $\simeq -2.0$~Nm on the shoulder.
In this way, the end-effector is not lowered excessively to avoid the unrepeatable situation where the weight hits the ground frequently, while horizontal dynamics would vary certainly.
At \textit{Interaction} scenario, the current position of the end-effector is measured through the forward kinematics, which is set to be the target position after passing through a low-pass filter.
Note that, in order to show the performance improvements by training, the impedance parameters fixed to their initial values are also tested.

%%%%%%%%%%%%%%%%%%%%%%%%%%%%%%%%%%%%%%%%%%%%%%%%%%%%%%%
\subsection{Learning tendency}

%Figure
\begin{figure}[tb]
    \centering
    \includegraphics[keepaspectratio=true,width=0.96\linewidth]{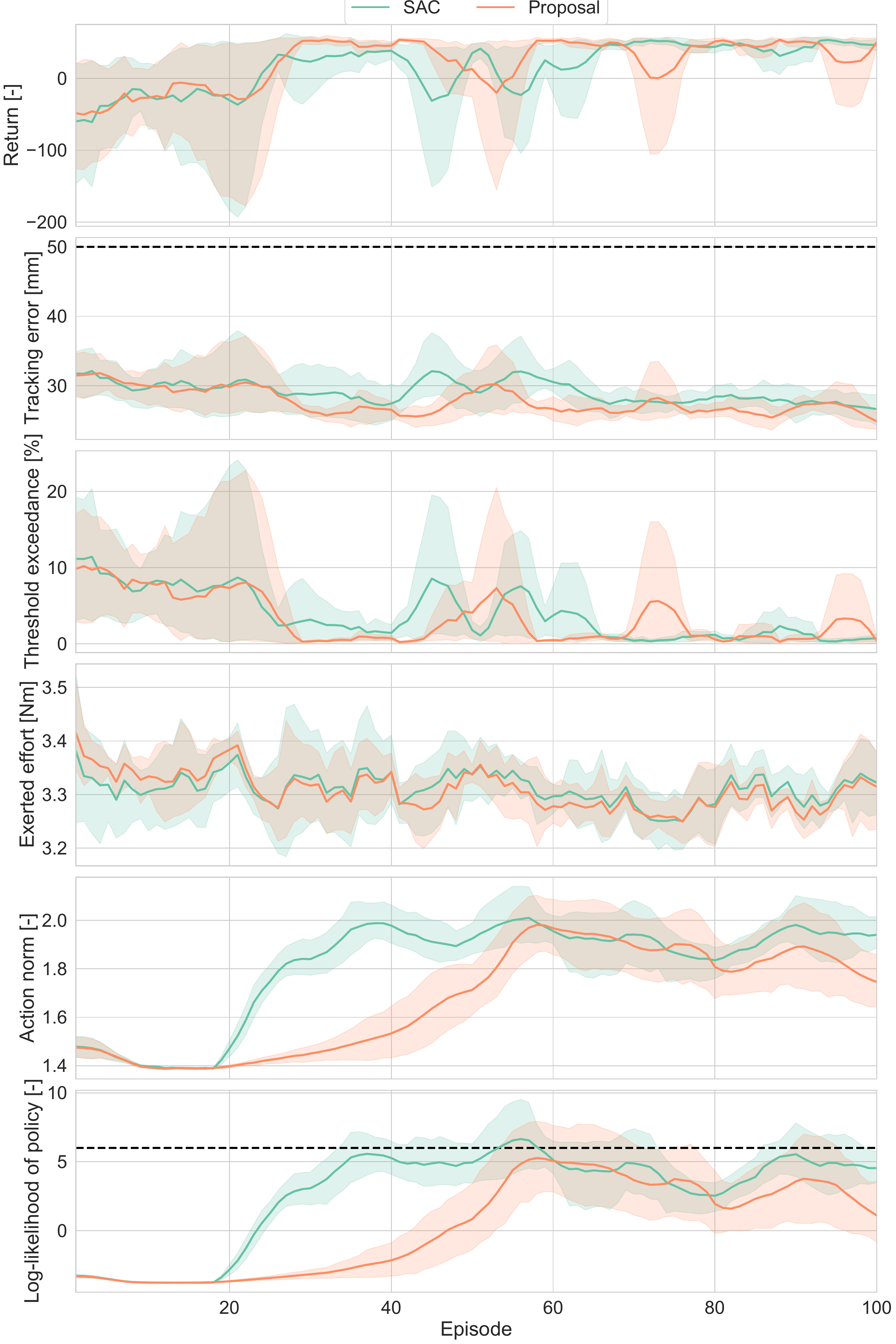}
    \caption{Learning curves on the variable impedance task}
    \label{fig:hebi_learn}
\end{figure}

The learning curves of the evaluated criteria are depicted in Fig.~\ref{fig:hebi_learn}.
Here, ``SAC'' shows the conventional SAC and ``Proposal'' shows the proposed method with $\mathcal{H}^{\ast} = \overline{\mathcal{H}} - 2|\mathcal{A}|$.
Although the proposed method was able to reduce the tracking error slightly, their returns were mostly similar during learning, as confirmed in the above simulations.
On the other hand, the policy entropy of the proposed method was clearly larger than that of the conventional method (i.e. the log-likelihood of the policy is smaller), indicating that the proposed method handled the inequality constraint properly even in the real-robot experiment.
Correspondingly, the action norm was smaller, making the policy more conservative.
Since it yields the smooth updates of the impedance parameters, it will play an important role for the stable operation of this system as below.

%%%%%%%%%%%%%%%%%%%%%%%%%%%%%%%%%%%%%%%%%%%%%%%%%%%%%%%
\subsection{Test results}

Table~\ref{tab:result} summarized the results of the above three scenarios, examples of which are in the attached video.
First, both SAC and Proposal outperformed Fixed on Vanilla scenario.
Looking at the mean of Return alone, a large difference between SAC and Proposal could not be found, but looking at SD, the performance was clearly stabilized by Proposal.
This is because Proposal acquired the skill not to exceed the threshold while allowing for some margin of error, implied in Tracking error and Threshold exceedance.
Note that a slight increase in Exerted effort after learning indicates that additional torques were needed to improve the tracking accuracy.

Second, on Weight scenario, the performance forceibly decreased with the change in dynamics with the grasped weight.
In such a situation, Fixed achieved the best performance since the other optimized impedance parameters caused resonance in $y$-axis motion on the rare occasion.
Nevertheless, Proposal caused it less frequently than SAC and maintained the performance close to Fixed (in fact, it outperformed Fixed when evaluated with the median).

Finally, on Interaction scenario, Fixed failed to track the human motions despite the torques were exerted uselessly.
In contrast, Proposal was able to maintain relatively good performance and win best in all the criteria.
In particular, Exerted effort, which in other scenarios was allowed to increase slightly for the sake of tracking accuracy, was
less increased than the others despite both Tracking error and Threshold exceedance were also best.
This suggests that Proposal did not make the impedance parameters too rigid in order to improve tracking accuracy, but rather optimized appropriately depending on situations.
In other words, Proposal only stiffened when the tracking error was large and was likely to exceed the threshold, and otherwise behaved softly, which enabled flexible tracking with little effort in response to external forces from the human.

Thus, the proposed method is confirmed that its policy is robust even to real-world perturbations, by correctly maximizing the policy entropy.
An interesting trend is that Log-likelihood of policy increased in the perturbed scenarios.
That is, the proposed method was able to deal with the inexperienced events appropriately because it learned the policy conservatively from similar experiences, while the conventional SAC resulted in incorrect actions due to insufficient robustness.

%Table
\begin{table*}[tb]
    \caption{Criteria of the variable impedance task: the best in each scenario is written bold.}
    \label{tab:result}
    \centering
    \begin{tabular}{l l | c c c c c c}
        \hline\hline
        Test & Method & \multicolumn{6}{c}{Criteria: Mean (SD)}
        \\
        && Return & Tracking error & Threshold exceedance & Exerted effort & Action norm & Log-likelihood of policy
        \\
        && [--] & [mm] & [\%] & [Nm] & [--] & [--]
        \\
        \hline
        Vanilla
        & Fixed
        & 36.44 & 30.18 & 1.52 & \textbf{3.36} & N/A & N/A
        \\
        && (14.28) & (1.77) & (1.32) & (0.16) & N/A & N/A
        \\
        & SAC
        & 44.24 & \textbf{25.21} & 0.70 & 3.40 & 1.84 & 5.61
        \\
        && (20.51) & (3.07) & (1.43) & (0.21) & (0.06) & (0.69)
        \\
        & Proposal
        & \textbf{46.68} & 25.69 & \textbf{0.56} & 3.39 & \textbf{1.59} & \textbf{2.71}
        \\
        && (14.75) & (2.85) & (0.60) & (0.21) & (0.16) & (2.57)
        \\
        \hline
        Weight
        & Fixed
        & \textbf{23.01} & 28.99 & \textbf{1.26} & \textbf{4.06} & N/A & N/A
        \\
        && (20.57) & (1.91) & (1.47) & (0.39) & N/A & N/A
        \\
        & SAC
        & 12.28 & 28.71 & 2.35 & 4.12 & 1.90 & 7.34
        \\
        && (45.71) & (4.21) & (4.11) & (0.50) & (0.08) & (2.62)
        \\
        & Proposal
        & 20.31 & \textbf{27.85} & 1.57 & 4.09 & \textbf{1.65} & \textbf{4.01}
        \\
        && (29.12) & (3.39) & (2.40) & (0.50) & (0.15) & (2.45)
        \\
        \hline
        Interaction
        & Fixed
        & -18.64 & 21.75 & 4.29 & 5.04 & N/A & N/A
        \\
        && (72.83) & (7.26) & (6.28) & (0.89) & N/A & N/A
        \\
        & SAC
        & 6.60 & 15.82 & 2.72 & 4.65 & 1.90 & 7.38
        \\
        && (71.61) & (9.42) & (5.39) & (1.33) & (0.08) & (2.64)
        \\
        & Proposal
        & \textbf{19.53} & \textbf{14.90} & \textbf{1.75} & \textbf{4.37} & \textbf{1.71} & \textbf{4.43}
        \\
        && (51.30) & (8.22) & (3.28) & (1.15) & (0.17) & (2.51)
        \\
        \hline\hline
    \end{tabular}
\end{table*}

%%%%%%%%%%%%%%%%%%%%%%%%%%%%%%%%%%%%%%%%%%%%%%%%%%%%%%%%%%%%%%%%%%%%%%%%%%%%%%%%
\section{Conclusion}

This paper addressed the new auto-tuning way of the temperature parameter in SAC.
Specifically, the previous method no longer maximizes the policy entropy, just constrains it to the lower bound (therefore, it is also called the target entropy).
This is because the way of auto-tuning is interpreted as the derivation from equality constraint.
To alleviate this issue, the desired inequality constraint to maximize the policy entropy was appropriately converted to the corresponding equality constraint using a state-dependent slack variable.
In addition, the slack variable was optimized to satisfy the equality constraint while minimizing its value as much as possible.
With the proposed improvement, the simulations on Mujoco and Pybullet verified the higher robustness to perturbation while regularizing the norm of action, which makes the policy conservative.
Finally, usefulness of the proposed method was exemplified by a real-robot experiment.
The proposed method exhibited adaptive behaviors even for physical human-robot interaction, which had no experience at all during training.

However, in the absence of perturbations, the proposed method was sometimes inferior to the conventional SAC.
A framework for adjusting the lower bound $\mathcal{H}^{\ast}$, as in the literature~\cite{xu2021target}, or a more theoretically valid lower bound should be investigated in the near future.

%%%%%%%%%%%%%%%%%%%%%%%%%%%%%%%%%%%%%%%%%%%%%%%%%%%%%%%%%%%%%%%%%%%%%%%%%%%%%%%%
\appendix

%%%%%%%%%%%%%%%%%%%%%%%%%%%%%%%%%%%%%%%%%%%%%%%%%%%%%%%%%%%%%%%%%%%%%%%%%%%%%%%%
\subsection{Implementation details}
\label{app:impl}

SAC is implemented in PyTorch~\cite{paszke2017automatic}, and the basic implementation follows the literature~\cite{kobayashi2022l2c2,kobayashi2022consolidated}.
During each episode, the tuple $(s, a, s^\prime, r)$ is obtained by interacting with the environment with the stochastically sampled action, storing it to the replay buffer $\mathcal{D}$ with the maximal size $N=102,400$.
At the end of the episode, up to half of the experiences are replayed uniformly at random by spliting them to the mini batchs with the maximal size $B=256$.
All the loss functions are calculated for each mini batch, and using the gradients of them, AdaTerm~\cite{ilboudo2022adaterm} with its default configuration optimizes all the network parameters defined below.

As for designing the functions to be optimized, the two value functions $Q_{1,2}$ are approximated by the independent neural networks, which have two hidden layers with 100 neuron for each.
The target networks for them, $\bar{Q}_{1,2}$, are updated by CAT-soft update~\cite{kobayashi2022consolidated} with its default configuration.
the policy $\pi$ is modeled by a diagonal multivariate student-t distribution, parameters of which (i.e. the location, the scale, and the degrees of freedom) are approximated by the shared neural networks except the output layer, which have the same architecture as $Q_{1,2}$.
The slack variable $\Delta$ is also approximated by the neural networks with the same architecture as $Q_{1,2}$ and $\pi$.
A squish function proposed in the literature~\cite{barron2021squareplus,kobayashi2023design} is employed as the activation functions for them.
In addition, RMSNorm~\cite{zhang2019root} (originally LayerNorm) is applied before the activation for the network normalization.

Note that SAC transforms the action generated from $\pi$, which is basically in real space, to the box-constrained one $a \in [-1, 1]^{|\mathcal{A}|}$.
This transformation is conventionally performed with Tanh function, which tends to concentrate the action near the boundary with large gradient.
For more stable gradient, an alternative transformation based on the literature~\cite{barron2021squareplus,kobayashi2023design}, which is also utilized for $\Delta$, is employed.

As one of the hyperparameters, the initial value of $\alpha$ is set to $1$.
The optimization of $\alpha$ (more specifically, its real-valued version $\tilde{\alpha}$) is conducted together with the networks.
$\epsilon$ in eq.~\eqref{eq:loss_delta} is simply tuned to $\epsilon=0.1|\mathcal{A}|$ because all terms in the equality constraint on $\Delta$ are proportional to $|\mathcal{A}|$.

%%%%%%%%%%%%%%%%%%%%%%%%%%%%%%%%%%%%%%%%%%%%%%%%%%%%%%%%%%%%%%%%%%%%%%%%%%%%%%%%
\subsection{Experimental configuration}
\label{app:robot}

For the variable impedance task, a 5-DoF manipulator as shown in Fig.~\ref{fig:hebi_arm} is used.
It consists of actuators (and frames) developed by HEBI Robotics, and each actuator is a series elastic actuator (SEA).
Therefore, The torque exerted can be calculated from the difference in rotation angles before and after its internal spring, and torque control is available accordingly.
In the experiment, only the official controllers provided by HEBI Robotics were used:
i.e. control strategy 4 to the target position and velocity; a gravity compensation; an offset to roughly compensate for the shoulder-mounted gas spring; and a task-space impedance controller with virtual spring and damper for each axis.
The PD gain is set lower than the default so that tracking errors are intentionally encountered, and the compensation by the impedance controller contributes to the performance.
Specifically, The P gains for the two actuators on a wrist are set to be 5; The P gains for the remaining three are set to be 20; The D gains for all are set to be 0.01.

For this manipulator, the following sine wave is given as the target trajectory of its end-effector.
\begin{align}
    p^\mathrm{tar}(t) = p^\mathrm{ini} \pm A \sin(\omega \pi t)
\end{align}
where $t$ denotes the elapsed time from the beginning of episode (with 20~ms control period) and $p^\mathrm{ini}$ is given by the forward kinematics for the center of each joint range as the initial position of the end-effector.
$A$ denotes the amplitude for each axis: $A_{xy} \in [0.05, 0.15]^2$ for horizontal plane and; $A_z \in [0.01, 0.05]$ for vertical direction.
$\omega \in [0.1, 0.9]^3$ for all axes are given as the axis-dependent frequency.
$A$, $\omega$, and the sign $\pm$ are uniformly randomized independently during each episode.
To track $p^\mathrm{tar}$, the target joint angle $q^\mathrm{tar}$ and velocity $\dot{q}^\mathrm{tar}$ are computed using the inverse kinematics~\cite{kobayashi2021mirror}, sending to each actuator as a command.

Under the above impedance control system for tracking the given trajectory, RL (i.e. SAC and the proposed method) optimizes the impedance parameters (i.e. three-dimensional virtual spring $k_p \in [0, 100]^3$ and damper $k_d \in [0, 10]^3$).
To this end, action space is designed as their variations (i.e. six dimensions in total), which are restricted to $\pm$5~\% of the maximum.
The following reward function is to track the target trajectory within an allowable error while minimizing effort.
\begin{align}
    r = \exp\left(- \sum_{i}|\tau_i| \right) -
    \begin{cases}
        1 & \|p^\mathrm{tar} - p^\mathrm{obs}\|_2 > 0.05
        \\
        0 & \mathrm{otherwise}
    \end{cases}
\end{align}
where $\tau$ denotes the efforts, $p^\mathrm{obs}$ is computed by the forward kinematics for the observed angles as the observed end-effector position.
To satisfy MDP with these configurations, the state space is given as a total of 32 dimensions consisting of the angle, angular velocity, effort, and temperature for each actuator; the observed and target positions of the end-effector; and the current impedance parameters.

%%%%%%%%%%%%%%%%%%%%%%%%%%%%%%%%%%%%%%%%%%%%%%%%%%%%%%%%%%%%%%%%%%%%%%%%%%%%%%%%
\section*{ACKNOWLEDGMENT}

This work was supported by JST, PRESTO Grant Number JPMJPR20C3, Japan.

%%%%%%%%%%%%%%%%%%%%%%%%%%%%%%%%%%%%%%%%%%%%%%%%%%%%%%%%%%%%%%%%%%%%%%%%%%%%%%%%
\bibliographystyle{IEEEtran}
{
\bibliography{biblio}
}

%%%%%%%%%%%%%%%%%%%%%%%%%%%%%%%%%%%%%%%%%%%%%%%%%%%%%%%%%%%%%%%%%%%%%%%%%%%%%%%%

\end{document}